\pdfoutput=1

\documentclass[10pt,twocolumn,letterpaper]{article}

\usepackage[pagenumbers]{cvpr} 

\usepackage{graphicx}
\usepackage{amsmath}
\usepackage{amssymb}
\usepackage{booktabs}

\usepackage{algorithm}
\usepackage{algpseudocode}

\usepackage{multirow} 
\usepackage{makecell}
\usepackage{color}
\usepackage{colortbl}
\usepackage[dvipsnames, svgnames, x11names]{xcolor}

\usepackage{enumitem}

\usepackage[pagebackref,breaklinks,colorlinks]{hyperref}

\usepackage[capitalize]{cleveref}
\crefname{section}{Sec.}{Secs.}
\Crefname{section}{Section}{Sections}
\Crefname{table}{Table}{Tables}
\crefname{table}{Tab.}{Tabs.}

\def\cvprPaperID{667} 
\def\confName{CVPR}
\def\confYear{2022}

\begin{document}

\title{Degradation-agnostic Correspondence from Resolution-asymmetric Stereo}

\author{Xihao Chen \\
Jiayong Peng
\and
Zhiwei Xiong\thanks{Corresponding author: zwxiong@ustc.edu.cn} \\
Yueyi Zhang
\and
Zhen Cheng \\
Zheng-Jun Zha
\and
University of Science and Technology of China
}

\maketitle

\begin{abstract}
   In this paper, we study the problem of stereo matching from a pair of images with different resolutions, \emph{e.g.}, those acquired with a tele-wide camera system.
   Due to the difficulty of obtaining ground-truth disparity labels in diverse real-world systems, we start from an unsupervised learning perspective.
   However, resolution asymmetry caused by unknown degradations between two views hinders the effectiveness of the generally assumed photometric consistency.
   To overcome this challenge, we propose to impose the consistency between two views in a feature space instead of the image space, named \textbf{feature-metric consistency}.
   Interestingly, we find that, although a stereo matching network trained with the photometric loss is not optimal, its feature extractor can produce degradation-agnostic and matching-specific features.
   These features can then be utilized to formulate a feature-metric loss to avoid the photometric inconsistency. 
   Moreover, we introduce a self-boosting strategy to optimize the feature extractor progressively, which further strengthens the feature-metric consistency.
   Experiments on both simulated datasets with various degradations and a self-collected real-world dataset validate the superior performance of the proposed method over existing solutions.
\end{abstract}

\section{Introduction}
\label{sec:intro}
Tele-wide camera systems consisting of two (or more) lenses with different focal lengths are widely deployed in smartphones nowadays.
This kind of systems usually generates a pair (or a set) of images with different resolutions at one shot, which enables a number of desirable applications, such as continuous optical zoom \cite{pan2016continuous} and image quality enhancement \cite{trinidad2019multi,wang2021dual,wang2021asymmetric}.
For these applications, correspondence estimation from resolution-asymmetric stereo images is a key step, which is typically conducted by conventional symmetric stereo matching algorithms (\emph{e.g.}, SGM \cite{hirschmuller2007stereo}) together with image upsampling \cite{pan2016continuous}.
However, this straightforward solution is vulnerable to the artifacts introduced by upsampling, especially when the upsampling scale is large.

\begin{figure}
  \centering
  \includegraphics[width=\columnwidth]{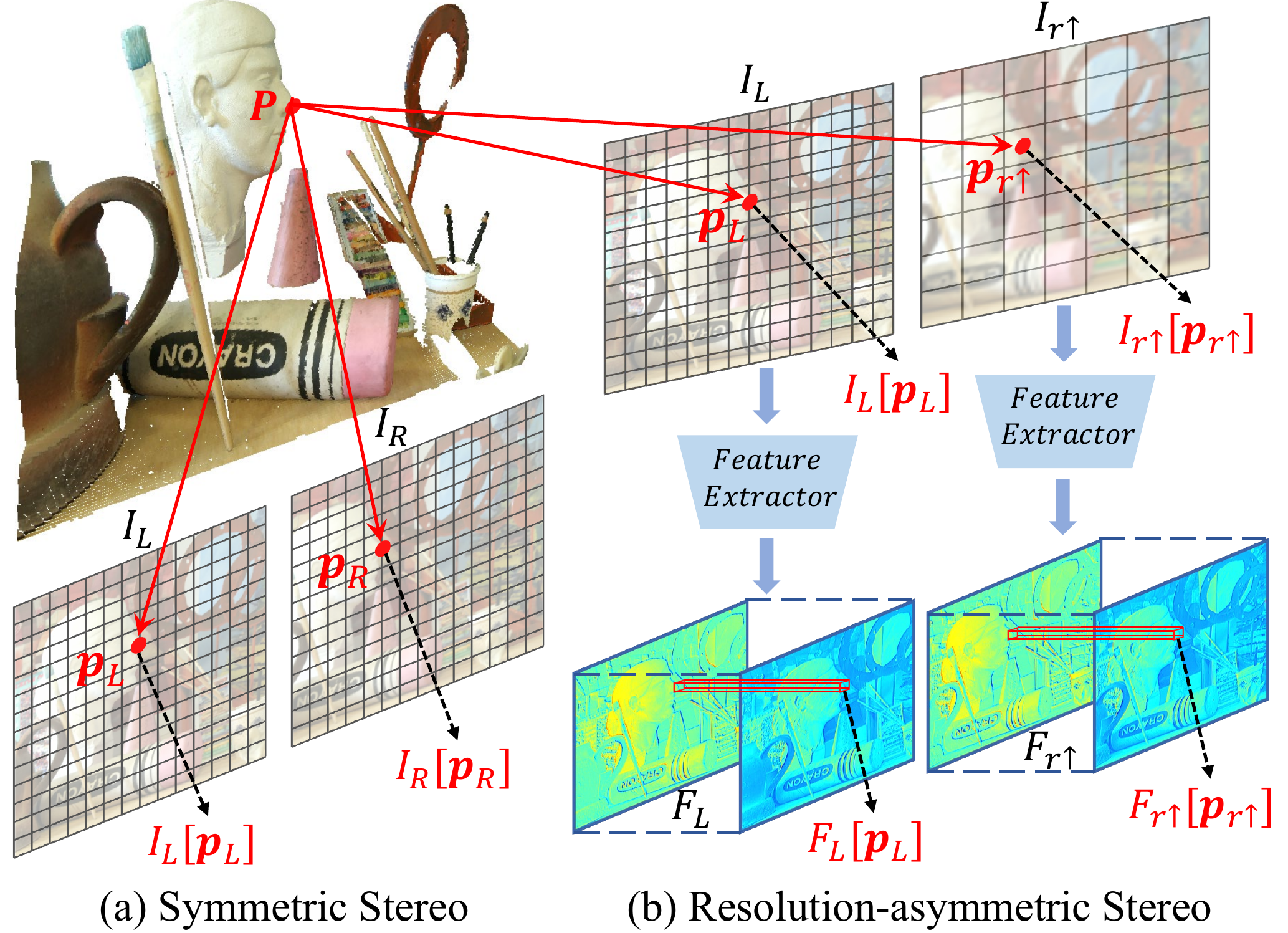}
  \vspace{-6mm}
  \caption{The common assumption of photometric consistency (\emph{i.e.}, $I_L[\textit{\textbf{p}}_L]$=$I_R[\textit{\textbf{p}}_R]$) in symmetric stereo is violated in resolution-asymmetric stereo (represented by dense and sparse grids). To avoid such photometric inconsistency (\emph{i.e.}, $I_L[\textit{\textbf{p}}_L]$$\neq$$I_{r\uparrow}[\textit{\textbf{p}}_{r\uparrow}]$), we establish the feature-metric consistency, which ensures that the pixels (\emph{e.g.}, $\textit{\textbf{p}}_L$ and $\textit{\textbf{p}}_{r\uparrow}$), recording the light rays (red arrows) emitted from the same scene point (\emph{e.g.}, $\textit{\textbf{P}}$), have the same feature representation (\emph{i.e.}, $F_L[\textit{\textbf{p}}_L]$=$F_{r\uparrow}[\textit{\textbf{p}}_{r\uparrow}]$).}
\label{fig:consistency}
\vspace{-4mm}
\end{figure}

Asymmetric stereo matching has been studied in literature under several specific contexts, \emph{e.g.}, radiometric variation \cite{hirschmuller2008evaluation} and modality difference \cite{zhi2018deep}.
In this paper, we focus on the resolution-asymmetric setting, which is practical yet has rarely been investigated explicitly.
As a recent related work, Liu \emph{et al.} propose a unified network for visually imbalanced stereo matching that addresses monocular blur and noise \cite{liu2020visually}.
Despite of its inspiring idea, this fully supervised approach requires not only the ground-truth disparity and the high-quality version of the degraded view as labels but also the explicit degradation \cite{bulat2018learn,chen2018image,xu2010two,huang2020real,wang2020deep} form to learn the parameters of the network, making it difficult to be applicable in diverse real-world systems where the supervision information is seldom available.
Therefore, we turn to the direction of unsupervised learning.

For unsupervised stereo matching, the most widely adopted assumption is photometric consistency \cite{zhong2017self}.
Under this assumption, the corresponding pixels in two views (\emph{e.g.}, $\textit{\textbf{p}}_L$ and $\textit{\textbf{p}}_R$ in Fig.~\ref{fig:consistency}(a)), which record the light rays emitted from the same scene point (\emph{e.g.}, \textit{\textbf{P}}), should have the same intensity or color (\emph{i.e.}, $I_L[\textit{\textbf{p}}_L]$=$I_R[\textit{\textbf{p}}_R]$).
Unfortunately, this assumption is violated for a resolution-asymmetric stereo pair, where the low-resolution (LR) view is degraded by an unknown downsampling kernel compared to the high-resolution (HR) view.
In other words, the corresponding pixels in the asymmetric stereo pair (\emph{e.g.}, $\textit{\textbf{p}}_L$ and $\textit{\textbf{p}}_{r\uparrow}$\footnote{$\uparrow$ denotes upsampling.} in Fig.~\ref{fig:consistency}(b)) may not have the same intensity or color (\emph{i.e.}, $I_L[\textit{\textbf{p}}_L]$$\neq$$I_{r\uparrow}[\textit{\textbf{p}}_{r\uparrow}]$).
Such photometric inconsistency will result in difficulties for correspondence learning.
A possible solution for remedy is to restore the LR view to an HR one by super-resolution (SR) techniques \cite{dong2014learning,zhang2018image,luo2020unfolding}.
However, existing SR methods are mostly degradation-specific and suffer from performance drops if the real degradation is different from the assumed one (for non-blind SR) or not inside the assumed range (for blind SR) \cite{chen2019camera,liu2021blind,zhang2019zoom,cai2019toward}.
Therefore, the effectiveness of SR methods to make up the photometric inconsistency will be hindered in practice.

To overcome the above challenge, we propose to solve resolution-asymmetric stereo matching from a new perspective by imposing the consistency of two views in a feature space instead of the image space, named feature-metric consistency.
Interestingly, we find that, although a stereo matching network trained with the photometric loss is not optimal, its feature extractor can produce degradation-agnostic (\emph{i.e.}, robustness to the degradation between $I_L$ and $I_{r\uparrow}$) and matching-specific features for corresponding asymmetric pixels (\emph{i.e.}, $F_L[\textit{\textbf{p}}_L]$=$F_{r\uparrow}[\textit{\textbf{p}}_{r\uparrow}]$ in Fig.~\ref{fig:consistency}(b)).
These features can then be utilized to formulate a feature-metric loss to avoid the photometric inconsistency.
Moreover, by finetuning the stereo matching network using the feature-metric loss, we can optimize the feature extractor to capture more consistent properties from the stereo pair, strengthening the feature-metric consistency.
To this end, we introduce a self-boosting strategy to optimize the feature extractor progressively.
Specifically, we use the feature extractor learned from the previous stage to form a new feature-metric loss for the current stage.
In this way, our method remains effective even for large degradations.

To quantitatively evaluate the performance of our method, we simulate four resolution-asymmetric stereo datasets, two from the widely used stereo datasets Middlebury \cite{hirschmuller2007evaluation} and KITTI2015 \cite{menze2015object} and two from the light field datasets Inria\_SLFD \cite{shi2019framework} and HCI \cite{honauer2016dataset} with a narrow baseline between two views which is closer to the configuration on smartphones.
The LR view is generated under various degradations from its original HR version.
To evaluate our method in real-world scenarios, we collect a resolution-asymmetric stereo dataset with the tele-wide camera system equipped on a Huawei P30 smartphone.
Experimental results on both simulated and real-world datasets demonstrate that our method outperforms existing as well as potential solutions by a large margin.

Contributions of this paper are summarized as follows:
\vspace{-2.5mm}
\begin{itemize}[leftmargin=*] \itemsep -2.5pt
\item The first unsupervised learning method for correspondence estimation from resolution-asymmetric stereo.
\item An effective and efficient realization of feature-metric consistency to avoid photometric inconsistency caused by unknown degradations.
\item A self-boosting strategy to strengthen feature-metric consistency by progressive loss update.
\item Distinct performance improvements over comparison methods on both simulated and real-world datasets.
\end{itemize}

\section{Related Work}
\noindent \textbf{Stereo Matching.} 
Stereo matching, symmetric by default, has been extensively studied as a classical computer vision task for decades \cite{scharstein2002taxonomy,hirschmuller2007stereo}.
Recently, deep learning based stereo matching methods have notably surpassed conventional algorithms.
According to whether or not ground-truth disparity maps are required as labels, these methods can be divided into supervised \cite{mayer2016large,kendall2017end,chang2018pyramid,cheng2020hierarchical,Li_VCIP_2021} and unsupervised \cite{zhong2017self,zhou2017unsupervised,wang2020parallax,aleotti2020reversing} categories.
In many real-world systems where the labels are not readily available, unsupervised methods enable learning without ground-truth information, most of which exploit the assumption of photometric consistency to formulate a photometric loss \cite{zhong2017self,wang2020parallax,tonioni2019real,zhou2019fast,Cheng_2021_CVPR,peng2020zero}.
However, this assumption will be violated when stereo images become asymmetric. 

\noindent \textbf{Asymmetric Stereo Matching.}
Several kinds of asymmetry have been considered in literature for stereo matching, including radiometric variation \cite{hirschmuller2008evaluation}, modality difference \cite{zhi2018deep}, and visual quality imbalance \cite{liu2020visually}.
To estimate correspondence from stereo images with radiometric variation, different robust matching costs are proposed, such as mutual information measure \cite{egnal2000mutual} and adaptive normalized cross-correlation \cite{heo2010robust}.
For cross-modal stereo \cite{zhi2018deep,Wang_2018_TCSVT,yao2019spectral}, images from two different modalities are normalized to a single one to make up the photometric inconsistency, \emph{e.g.}, through deep transformation networks \cite{zhi2018deep,liang2019unsupervised}.
Recently, stereo matching with visual imbalance (monocular blur and noise) is addressed by integrating a view synthesis network and a stereo reconstruction network, which requires the ground-truth disparity, the high-quality version of the degraded view, and the explicit degradation form for supervision \cite{liu2020visually}.
Resolution asymmetry can be regarded as a certain kind of visual imbalance, but such a supervised solution is difficult to be applicable in diverse real-world systems.

\noindent \textbf{Feature-metric Learning.}
For geometry tasks, there are several pioneering works to utilize deep features as the metric of unsupervised learning.
Specifically, Zhang \emph{et al.} improve the performance of monocular depth estimation by integrating the photometric loss and a feature-metric loss based on pre-trained features \cite{zhan2018unsupervised}.
Different from \cite{zhan2018unsupervised}, Shu \emph{et al.} learn customized features with an auto-encoder and two regularizing losses \cite{shu2020feature}, while Spencer \emph{et al.} learn features with the contrastive loss \cite{spencer2020defeat}.
For domain adaptation, Liu \emph{et al.} propose to penalize the matching error of a stereo network in the feature space of a domain translation network \cite{liu2020stereogan}.
Inspired by the above works, for the first time, we introduce the concept of feature-metric consistency to the new task of resolution-asymmetric stereo matching.

\section{Preliminary}
A pair of resolution-asymmetric stereo images consists of an HR view and an LR view.
Without loss of generality, we take the left view $I_L \in \mathbb{R}^{H \times W}$ as the HR view and the right view $I_r \in \mathbb{R}^{\frac{H}{s} \times \frac{W}{s}}$ as the LR view, where $s$ is an asymmetric factor.
To align their resolutions, $I_r$ is upsampled with a classical interpolation algorithm (\emph{e.g.}, bicubic), denoted as $I_{r\uparrow} \in \mathbb{R}^{H \times W}$.
Despite being upsampled, the high-frequency information in $I_{r\uparrow}$ is absent, and thus the stereo pair $I_L$ and $I_{r\uparrow}$ is still asymmetric.

\subsection{Learning with Photometric Consistency}
\label{sec_unsu_pm}
Given a stereo pair $I_L$ and $I_{r\uparrow}$ as input, an unsupervised stereo matching network $\Phi(\cdot;\theta)$ aims to predict a disparity map for $I_L$, denoted as $d_L=\Phi(I_L, I_{r\uparrow};\theta)$, under 
the assumption of photometric consistency between the corresponding pixels in two views (denoted as $\textit{\textbf{p}}_L$ and $\textit{\textbf{p}}_{r\uparrow}$), \emph{i.e.},
\vspace{-2.75mm}
\begin{equation}
\label{eq:ph_consistency}
      I_L[\textit{\textbf{p}}_L]=I_{r\uparrow}[\textit{\textbf{p}}_{r\uparrow}].
\vspace{-0.5mm}         
\end{equation}
If the disparity $d_L[\textit{\textbf{p}}_L]$ between $\textit{\textbf{p}}_L$ and $\textit{\textbf{p}}_{r\uparrow}$ is accurately estimated, $I_L[\textit{\textbf{p}}_L]$ in the left view can be well reconstructed by warping $I_{r\uparrow}[\textit{\textbf{p}}_{r\uparrow}]$ in the right view with this disparity as
\vspace{-1.25mm}
\begin{equation}
   \label{eq:warp}
   I_{r\uparrow \rightarrow L}[\textit{\textbf{p}}_L] = I_{r\uparrow}[\textit{\textbf{p}}_L-d_L[\textit{\textbf{p}}_L]].   
   \vspace{-1.25mm} 
\end{equation}
Therefore, the photometric loss is formulated as the reconstruction error between $I_L$ and its reconstructed version $I_{r\uparrow \rightarrow L}$, typically in the form of a weighted combination of $L_1$ and SSIM distance, \emph{i.e.},
\vspace{-1.5mm}
\begin{equation}
   \label{eq:lpm}
   \mathcal{L}_{pm} = \left\|I_L-I_{r\uparrow \rightarrow L}\right\|_{1} + \alpha (1-{\rm SSIM}(I_L,I_{r\uparrow \rightarrow L})),
   \vspace{-1.25mm}
\end{equation}
where $\alpha$ is a weighting factor.

\subsection{Challenge and Motivation}
\label{sec_challenge}
Intuitively, resolution asymmetry challenges unsupervised stereo matching in twofold:
(i) It may be more difficult for the feature extractor of the network to extract symmetric features from the asymmetric input, and
(ii) the photometric loss may lose efficacy as Eq.~(\ref{eq:ph_consistency}) does not hold for asymmetric stereo.
We conduct a series of experiments to verify the influence of these two factors.
In the experiments, the ground-truth HR version $I_R$ of the right view is assumed to be available.
Therefore, we can control the symmetry (Sym) or asymmetry (Asy) of the images input to the feature extractor to ablate factor (i) and control the symmetry or asymmetry of the images used to compute the photometric loss to ablate factor (ii).

\begin{figure}
   \centering
   \includegraphics[width=0.98\columnwidth]{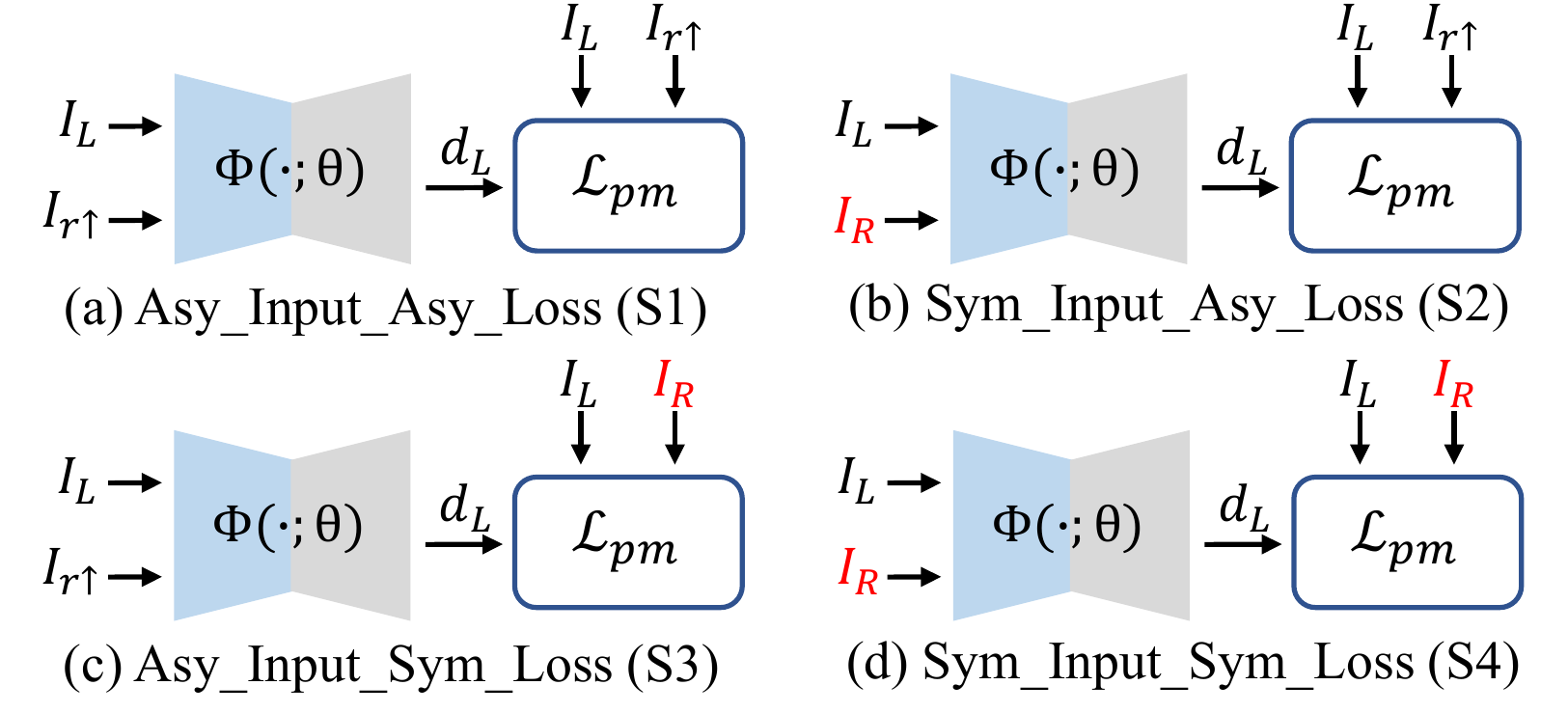}
   \vspace{-3mm}
   \caption{Illustration of four unsupervised stereo matching settings trained with photometric consistency.}
 \label{fig:four_settings}
 \vspace{-2mm}
 \end{figure}
 
\begin{table}[!t]
   \caption{3PE (\%) results of different unsupervised stereo matching settings on the Inria\_SLFD dataset.}
   \vspace{-3mm}
  \label{table:benchmark}
   \setlength{\tabcolsep}{8pt}
   \renewcommand{\arraystretch}{1.025}
   \centering
   \resizebox{0.95\columnwidth}{!}{
   \begin{tabular}{ccccc}
      \bottomrule[1.2pt]
      \multirow{2}{*}{Setting} & \multicolumn{4}{c}{Asymmetric Factor $s$} \\
      \cline{2-5}
         & 2 & 4 & 6 & 8  \\
      \hline
      Asy\_Input\_Asy\_Loss (S1) & 7.28 & 12.56 & 22.72 & 27.93  \\
      Sym\_Input\_Asy\_Loss (S2) & 7.22 & 10.01 & 16.31 & 21.93 \\
      Asy\_Input\_Sym\_Loss (S3) & 6.38 & 6.39 & 6.58 & 7.52 \\
      Sym\_Input\_Sym\_Loss (S4) & \multicolumn{4}{c}{6.32}  \\
      \toprule[1.2pt]
   \end{tabular}
   }
   \vspace{-3mm}
\end{table}

As shown in Fig. \ref{fig:four_settings}, a total of four settings of unsupervised stereo matching are evaluated, among which only the first one (S1) can be achieved in practice and the rest ones (S2, S3, and S4) can be regarded as ``ideal cases'' since the HR right view is used.
We select two views of each scene from the Inria\_SLFD dataset \cite{shi2019framework} as the HR left and right views, \emph{i.e.}, $I_L$ and $I_R$.
The LR right view $I_r$ is simulated from $I_R$ with bicubic downsampling under four asymmetric factors ($s$ = 2, 4, 6, 8).
We adopt the popular PSMNet \cite{chang2018pyramid} as the backbone network $\Phi(\cdot;\theta)$ and the photometric loss $\mathcal{L}_{pm}$ is computed following Eq.~(\ref{eq:lpm}) with $\alpha=3$.
A standard stereo matching metric, 3-Pixel-Error (3PE) \cite{menze2015object}, is used to evaluate the performance of different settings
(see more implementation details in Sec. \ref{subsec_experiment}).

As can be seen from Table~\ref{table:benchmark}, when the images input to the feature extractor change from asymmetric to symmetric (S1 to S2), the performance improvements are rather limited (\emph{e.g.}, 2.55\% when $s$ = 4).
In contrast, when the images used to compute the photometric loss change from asymmetric to symmetric (S1 to S3), the results see a large improvement (\emph{e.g.}, 6.24\% when $s$ = 4), which are even close to the upper bound (S4).
It is worth emphasizing that, for S1 and S3, the disparity maps used to warp the right view are from the same input and the same network.
This phenomenon can be observed under all asymmetric factors.
It clearly demonstrates that, for resolution-asymmetric stereo matching, the asymmetry during loss computation has a dominant influence rather than the asymmetry of the input.

A possible solution to make up the photometric inconsistency is to restore the LR right view $I_{r\uparrow}$ through SR techniques to approach $I_R$.
However, for diverse real-world systems, neither the realistic pair of ($I_{r\uparrow}$, $I_R$) nor the explicit degradation from $I_R$ to $I_{r\uparrow}$ can be easily available to train SR models.
Therefore, this solution may perform decently on properly simulated data but lose efficacy in practice.
In view of the results in Table~\ref{table:benchmark}, we propose to conquer the challenge of ``asymmetric loss'' from a new perspective, by projecting $I_L$ and $I_{r\uparrow}$ to a feature space that is agnostic to degradation and specific for matching. 
On the one hand, a degradation-agnostic space can establish another kind of consistency (\emph{i.e.}, \textbf{feature-metric consistency}) to avoid the photometric inconsistency.
On the other hand, a matching-specific space can assign different values to pixels belonging to different scene points and thus is suitable for penalizing incorrect matchings.
Now the remaining question is: how to learn the desirable feature space?

\section{Resolution-asymmetric Stereo Matching}
\subsection{Feature Space Investigation}
\label{sec_feature_selec}
Recalling the results in Table~\ref{table:benchmark}, it reveals that the feature extractor of a stereo matching network trained under the setting of S3 performs well in extracting symmetric features from the asymmetric input.
Although S3 is not attainable in practice, it suggests a potential substitute, \emph{i.e.}, the feature extractor of S1 that takes the same input, for obtaining the desirable feature space.
To validate this speculation, we conduct another series of experiments. 
Besides S1, we investigate two other representative feature spaces used for geometry tasks:
1) a feature network trained with the Contrastive Loss (denoted as CL) as in \cite{spencer2020defeat},
and 2) the encoder of an Auto-Encoder network (denoted as AE) as in \cite{shu2020feature}. Details of these two networks are provided in the supplement.
All the above networks are pre-trained on the Inria\_SLFD dataset with $s$ = 4.
Additionally, we also include the original image space for comparison.

We evaluate the degradation-agnostic property of different spaces by computing the PSNR metric between the feature maps extracted from $I_R$ and its degraded version $I_{r\uparrow}$ by the corresponding networks.
The PSNR in the image space is computed based on pixel intensities.
Note that the values in different spaces are normalized to $[0,1]$ to make the comparison of PSNR results meaningful.
On the other hand, for the matching-specific property, we perform matching between the feature maps extracted from $I_L$ and $I_{r\uparrow}$ directly in different spaces.
Specifically, we formulate a matching cost by computing the euclidean distance of two feature vectors at a given disparity.
Then a disparity map is obtained by selecting the minimal matching cost at each location following the Winner-Takes-All strategy.
For the image space, we perform matching on a 5$\times$5 patch basis.
The 3PE metric is then used to evaluate the obtained disparity map.

\begin{table}[!t]
   \caption{Evaluation of the degradation-agnostic property in PSNR (dB) and the matching-specific property in 3PE (\%) of different spaces on the Inria\_SLFD dataset.}
   \label{tab:observation}
   \vspace{-3mm}
   \setlength{\tabcolsep}{12pt}
   \renewcommand{\arraystretch}{1.0}
   \centering
   \resizebox{0.9\columnwidth}{!}{
   \centering
   \begin{tabular}{ccccc}
      \toprule[1pt]
       & Image & CL & AE & S1  \\
      \hline 
      PSNR $\uparrow$  & 24.65 & 44.18 & 23.23 & 28.00 \\ 
      3PE $\downarrow$  & 55.3 & 68.90 & 39.22 & 20.91 \\
      \bottomrule[1pt]
   \end{tabular}
   }
   \vspace{-2mm}
\end{table}

Table~\ref{tab:observation} gives the PSNR and 3PE results of different spaces.
Although CL presents the highest PSNR value, it performs worst in terms of 3PE.
In other words, CL is most degradation-agnostic but least matching-specific, which can be attributed to the blur feature maps extracted by the feature network.
AE can learn relatively discriminative features for matching thanks to the regularization losses, but it does not impose the consistency between the feature maps of $I_R$ and $I_{r\uparrow}$, resulting in the lowest PSNR value.
Compared with the image space where the photometric consistency is damaged by the degradation, the feature space of S1 can assign more consistent features for $I_R$ and $I_{r\uparrow}$ (with a notably higher PSNR value).
Meanwhile, this feature space is more discriminative for performing matching between $I_L$ and $I_{r\uparrow}$ than others (with the best 3PE result).
In conclusion, we verify that the feature extractor of a stereo matching network can approach the desirable feature space, even trained with the ``asymmetric loss''. More analysis on this part and the visualization of different feature maps are provided in the supplement.

\begin{figure*}
   \centering
   \includegraphics[width=\linewidth]{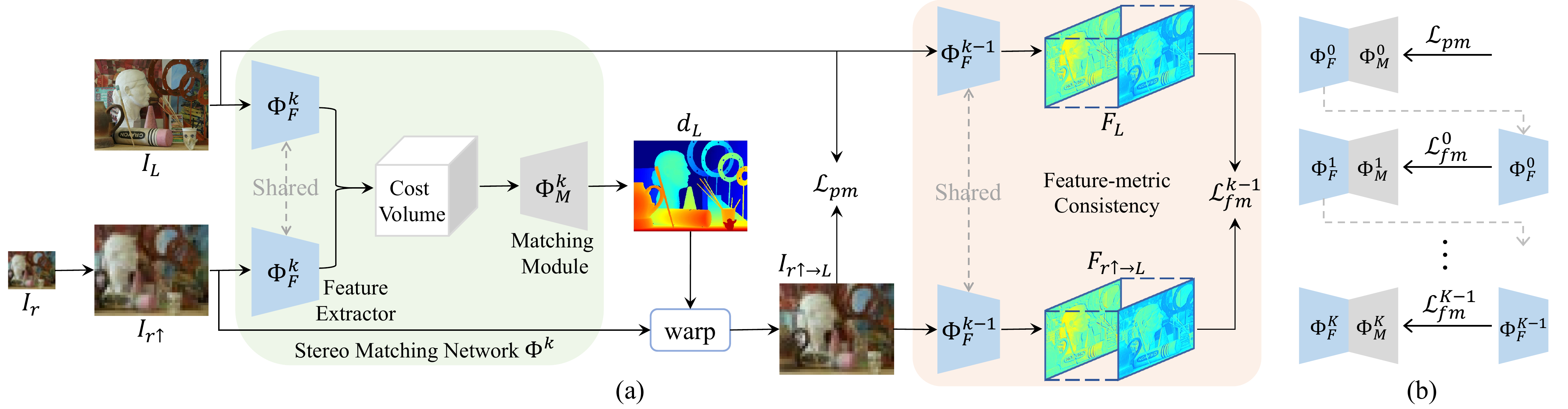}
   \vspace{-7mm}
   \caption{The proposed resolution-asymmetric stereo matching method. (a) The feature extractor $\Phi_F$ of a stereo matching network $\Phi$ is utilized to establish the feature-metric consistency and formulate a feature-metric loss $\mathcal{L}_{fm}$. (b) A self-boosting strategy is introduced to progressively strengthen the feature-metric consistency and continuously boost the network $\Phi$. Specifically, $\Phi_F^{k-1}$ obtained in the previous stage formulates $\mathcal{L}_{fm}^{k-1}$ to train $\Phi^k$ in the current stage, and the initial  $\Phi_F^0$ is trained with the photometric loss $\mathcal{L}_{pm}$.}
   \label{fig:framework}
   \vspace{-3mm}
 \end{figure*}

\subsection{Learning with Feature-metric Consistency}
Fig.~\ref{fig:framework} illustrates our proposed method for resolution-asymmetric stereo matching, which follows the typical pipeline of unsupervised learning as described in Sec.~\ref{sec_unsu_pm}.
Note that the focus of this work is \textit{not} to design a specific stereo matching network but to realize the feature-metric consistency to avoid the photometric inconsistency.
Therefore, we adopt the popular PSMNet \cite{chang2018pyramid} as the backbone of the stereo matching network, which can be readily replaced by other embodiments (see Sec.~\ref{sec:model_archi} for the embodiment of iResNet \cite{liang2018learning}). 

As illustrated in Fig.~\ref{fig:framework}(a), the stereo matching network $\Phi(\cdot;\theta_F,\theta_M)$ is comprised of a feature extractor $\Phi_F(\cdot;\theta_F)$ and a matching module $\Phi_M(\cdot;\theta_M)$.
Given a stereo pair $I_L$ and $I_{r\uparrow}$, $\Phi_F(\cdot;\theta_F)$ extracts degradation-agnostic and matching-specific features $F_L=\Phi_{F}(I_L;\theta_F)$ and $F_{r\uparrow}=\Phi_{F}(I_{r\uparrow};\theta_F)$, which are supposed to be consistent at corresponding asymmetric pixels ($\textit{\textbf{p}}_L$ and $\textit{\textbf{p}}_{r\uparrow}$), \emph{i.e.},
\vspace{-1mm}
\begin{equation}
   F_L[\textit{\textbf{p}}_L]=F_{r\uparrow}[\textit{\textbf{p}}_{r\uparrow}].
   \vspace{-1mm}
\end{equation}
Then, the features $F_L$ and $F_{r\uparrow}$ are concatenated into a cost volume that is regularized by $\Phi_M(\cdot;\theta_M)$ to regress a disparity map $d_L$.

According to the investigation in Sec.~\ref{sec_feature_selec}, we propose to use the feature extractor of the stereo matching network itself to produce the desirable feature space for computing a feature-metric loss.
Specifically, after obtaining the warped left view $I_{r\uparrow \rightarrow L}$ with $d_L$, the feature extractor $\Phi_F(\cdot;\theta_F)$ is used to project $I_L$ and $I_{r\uparrow \rightarrow L}$ to the feature space, producing $F_L$ and $F_{r\uparrow \rightarrow L} = \Phi_F(I_{r\uparrow \rightarrow L};\theta_F)$.
Since $F_L$ should be well reconstructed by $F_{r\uparrow \rightarrow L}$ if $d_L$ is estimated accurately, we can formulate the feature-metric loss with the reconstruction error similar to the photometric loss $\mathcal{L}_{pm}$ in Eq.~(\ref{eq:lpm}), denoted as
\vspace{-1mm}
\begin{equation}
   \label{eq:dif}
   \mathcal{L}_{fm} = \left\|F_L-F_{r\uparrow \rightarrow L}\right\|_{1} + \alpha (1-{\rm SSIM}(F_L,F_{r\uparrow \rightarrow L})).
\end{equation}

\subsection{Self-boosting Strategy}
\vspace{-0.75mm}
As demonstrated in Sec.~\ref{sec_feature_selec}, even when the stereo matching network $\Phi(\cdot;\theta_F,\theta_M)$ trained with the photometric loss $\mathcal{L}_{pm}$, its feature extractor $\Phi_F(\cdot;\theta_F)$ can approach the desirable feature space.
Nonetheless, when the network is trained by a more accurate loss (\emph{e.g.}, $\mathcal{L}_{fm}$), the corresponding $\Phi_F(\cdot;\theta_F)$ extracts more degradation-agnostic and matching-specific features, which can be utilized to strengthen the feature-metric consistency and formulate a better $\mathcal{L}_{fm}$.
In return, a better $\mathcal{L}_{fm}$ can further boost $\Phi(\cdot;\theta_F,\theta_M)$.
To this end, we propose a self-boosting strategy to progressively optimize the feature extractor and continuously boost the network.

Fig.~\ref{fig:framework}(b) illustrates the training process of our method.
Given a resolution-asymmetric stereo dataset, we first use $\mathcal{L}_{pm}$ to train a stereo matching network $\Phi(\cdot;\theta_F^0,\theta_M^0)$ (short as $\Phi^0$), whose feature extractor $\Phi_F^0$ formulates a feature-metric loss $\mathcal{L}_{fm}^0$.
Then, $\mathcal{L}_{fm}^0$ is utilized to finetune a new stereo matching network $\Phi^1$ which is initialized as $\Phi^0$.
During the finetuning of $\Phi^1$, the feature extractor for computing $\mathcal{L}_{fm}^0$ is fixed.
After finetuning, a boosted feature extractor $\Phi_F^1$ formulates a better feature-metric loss $\mathcal{L}_{fm}^1$, which is utilized in the next training stage.
Following this way, we iteratively finetune $\Phi^k$ with the progressively boosted $\mathcal{L}_{fm}^{k-1}$ ($k=1$,...,$K$).
Note that we only formulate a new training loss $\mathcal{L}_{fm}^k$ when the network $\Phi^k$ converges with respect to $\mathcal{L}_{fm}^{k-1}$, since altering the loss space frequently could make the training process unstable.
With this self-boosting strategy, we can obtain continuously optimized networks with progressively strengthened feature-metric consistency.
The detailed algorithm is provided in the supplement.

\begin{table}[!t]
   \caption{Validation of the self-boosting strategy on the Inria\_SLFD dataset. The 3PE (\%) metric is evaluated.}
   \label{tab:ablation_sbls}
   \vspace{-3mm}
   \setlength{\tabcolsep}{11pt}
   \renewcommand{\arraystretch}{1.}
   \centering
   \resizebox{0.89\columnwidth}{!}{
   \centering
   \begin{tabular}{ccccc}
      \toprule[1.1pt] 
      \multirow{2}{*}{\makecell[c]{Asymmetric\\Factor $s$}} & \multicolumn{4}{c}{Stage Number $k$} \\
      \cline{2-5}
       & 0 & 1 & 2 & 3 \\
      \hline  
      4 & 12.56 & 9.22 & 7.80 & 7.70 \\ 
      6 & 21.47 & 13.92 & 10.54 & 9.88 \\ 
      8 & 27.93 & 18.47 & 14.30 & 13.30 \\ 
      \bottomrule[1.1pt] 
   \end{tabular}
   }
   \vspace{-2mm}
\end{table}

\begin{table*}[!t]
   \caption{Comparison of different methods on four resolution-asymmetric stereo datasets simulated with an asymmetric factor of 4 and under various degradations. The 3PE (\%) / EPE metrics are evaluated. For SR solutions, the results marked gray denote their assumed degradations are inconsistent with the actual ones. The best results are highlighted with bold fonts.
   }
   \vspace{-3mm}
   \label{tab:overall_result}
   \setlength{\tabcolsep}{4pt}
   \renewcommand{\arraystretch}{1.1}
   \centering
   \resizebox{\linewidth}{!}{
   \begin{tabular}{|c|ccccc|ccccc|}
      \hline
      \multirow{2}{*}{Method} & \multicolumn{5}{c|}{Inria\_SLFD} & \multicolumn{5}{c|}{Middlebury} \\
      \cline{2-11}
      & BIC & IG & AG & IG\_JPEG & AG\_JPEG  & BIC & IG & AG & IG\_JPEG & AG\_JPEG  \\
      \hline
      SGM & 12.41/1.849 & 16.88/2.316 & 14.85/2.127 & 16.93/2.318 & 14.94/2.134 & 8.87/1.535 & 11.70/1.822 & 10.35/1.696 & 11.94/1.844 & 10.60/1.713 \\
      BaseNet                         & 12.56/1.680 & 16.75/2.158 & 15.27/1.996 & 16.42/2.029 & 13.40/1.844 & 8.72/1.363 & 9.50/1.482 & 8.89/1.416 & 10.27/1.613 & 8.61/1.414 \\ 
      RCAN+BaseNet & 8.89/1.287 & \cellcolor{Gainsboro}14.40/1.842 & \cellcolor{Gainsboro}12.34/1.604 & \cellcolor{Gainsboro}13.94/1.796 & \cellcolor{Gainsboro}12.01/1.612 & 6.76/1.189 & \cellcolor{Gainsboro}9.14/1.425 & \cellcolor{Gainsboro}7.86/1.287 & \cellcolor{Gainsboro}9.46/1.442 & \cellcolor{Gainsboro}8.72/1.381 \\
      DAN+BaseNet & 9.91/1.374 & 10.99/1.464 & 10.51/1.464 &  \cellcolor{Gainsboro}12.97/1.785 &  \cellcolor{Gainsboro}11.56/1.583 & 6.90/1.187     & 6.70/1.204 & 7.18/1.231 &  \cellcolor{Gainsboro}8.95/1.450 &  \cellcolor{Gainsboro}8.35/1.344 \\ 
      BaseNet+CL & 12.97/1.700 & 16.74/2.186 & 17.36/2.089 & 17.46/2.236 & 18.08/2.263 & 8.13/1.430 & 11.25/1.649 & 11.62/1.679 & 12.45/1.817 & 10.06/1.631 \\
      BaseNet+AE & 10.47/1.478 & 15.17/1.984 & 13.63/1.840 & 15.14/1.947 & 14.29/1.927 & 6.95/1.244 & 8.47/1.384 & 7.80/1.356 & 9.47/1.459 & 8.06/1.358 \\
      Ours & \textbf{7.70}/\textbf{1.148} & \textbf{9.01}/\textbf{1.337} & \textbf{8.44}/\textbf{1.249} & \textbf{9.65}/\textbf{1.418} & \textbf{8.47}/\textbf{1.288} & \textbf{5.78}/\textbf{1.088} & \textbf{6.52}/\textbf{1.178} & \textbf{6.38}/\textbf{1.172} & \textbf{7.04}/\textbf{1.204} & \textbf{7.05}/\textbf{1.203} \\ 
      \hline
         & \multicolumn{5}{c|}{HCI} & \multicolumn{5}{c|}{KITTI2015} \\
      \cline{2-11}
      \hline
      SGM & 7.04/1.093 & 9.85/1.426 & 8.50/1.273 & 10.02/1.425 & 8.62/1.278  & 30.71/4.001 & 38.90/5.043 & 36.01/4.659 & 39.04/5.040 & 36.14/4.660 \\ 
      BaseNet & 5.95/0.891 & 9.91/1.213 & 8.03/1.068 & 9.82/1.189 & 7.88/1.083 & 11.32/2.014 & 17.37/2.531 & 13.85/2.243 & 15.31/2.311 & 14.66/2.314 \\
      RCAN+BaseNet & 5.34/0.717 &  \cellcolor{Gainsboro}7.23/0.994 &  \cellcolor{Gainsboro}6.62/0.893 &  \cellcolor{Gainsboro}8.18/1.054 &  \cellcolor{Gainsboro}7.70/1.052 & 9.94/1.846 &  \cellcolor{Gainsboro}13.30/2.141 &  \cellcolor{Gainsboro}10.98/1.937 &  \cellcolor{Gainsboro}13.31/2.162 &  \cellcolor{Gainsboro}11.95/2.052 \\ 
      DAN+BaseNet & 5.48/0.715 & 5.32/0.781 & 6.23/0.830 &  \cellcolor{Gainsboro}7.86/0.988 &  \cellcolor{Gainsboro}6.56/0.984 & 10.06/1.938 & 10.31/\textbf{1.856} & 10.31/1.892 &  \cellcolor{Gainsboro}12.71/2.089 &  \cellcolor{Gainsboro}11.39/1.973 \\ 
      BaseNet+CL & 7.80/0.990 & 8.68/1.124 & 8.74/1.144 & 9.35/1.223 & 8.29/1.137 & 17.04/2.472 & 31.03/3.388 & 20.00/2.676 & 21.12/2.733 & 22.30/2.902 \\
      BaseNet+AE & 5.13/0.818 & 6.30/1.018 & 5.51/0.922 & 7.15/1.079 & 5.56/0.973 & 10.53/1.911 & 15.25/2.316 & 13.25/2.102 & 15.05/2.219 & 13.42/2.122 \\
      Ours & \textbf{4.08}/\textbf{0.637} & \textbf{4.56}/\textbf{0.701} & \textbf{4.21}/\textbf{0.670} & \textbf{4.58}/\textbf{0.719} & \textbf{4.35}/\textbf{0.709} & \textbf{8.66}/\textbf{1.801} & \textbf{10.08}/1.901 & \textbf{9.70}/\textbf{1.848} & \textbf{10.62}/\textbf{1.948} & \textbf{9.82}/\textbf{1.874} \\
      \hline
   \end{tabular}
   }
   \vspace{-3mm}
\end{table*}

To validate the proposed strategy, we evaluate the performance of the stereo matching networks at different stages on the Inria\_SLFD dataset with $s=4$.
As can be seen in Table~\ref{tab:ablation_sbls}, the network is progressively improved with the increase of stages.
It reflects that the feature extractor used in the next stage is boosted and the feature-metric consistency is strengthened.
Moreover, with such a strategy, our method remains effective for large degradations.
We validate this claim with two larger asymmetric factors ($s$ = 6, 8).
As shown in Table~\ref{tab:ablation_sbls}, the performance of the initial network ($k=0$) significantly deteriorates due to the more severe photometric inconsistency when the asymmetric factor increases.
However, the network finally reaches decent performance, thanks to the self-boosting strategy.

\section{Experiments on Simulated Datasets}
\label{subsec_experiment}
\subsection{Datasets and Evaluation Metrics}
To quantitatively evaluate the performance of our method, we simulate four resolution-asymmetric stereo datasets, two from the widely used stereo datasets Middlebury \cite{hirschmuller2007evaluation} and KITTI2015 \cite{menze2015object} and two from the light field datasets Inria\_SLFD \cite{shi2019framework} and HCI \cite{honauer2016dataset} with a narrow baseline between two views which is closer to the configuration on smartphones.
To mimic the diverse degradations in real-world systems, we perform five different degradation operations to synthesize the LR view, including bicubic downsampling (BIC), Isotropic/Anisotropic Gaussian kernel downsampling (IG/AG), and Isotropic/Anisotropic Gaussian kernel downsampling with JPEG compression (IG\_JPEG/AG\_JPEG).
Details of training/testing division of each dataset and generation of different Gaussian kernels are provided in the supplement.
For performance evaluation, we adopt two standard metrics for stereo matching, 3-Pixel-Error (3PE) \cite{menze2015object} and End-Point-Error (EPE) \cite{mayer2016large}.
3PE is the percentage of the predicted disparities whose errors are more than 3 pixels and 5\% of their ground-truth disparities, while EPE is the average absolute difference between the estimated and ground-truth disparities.

\subsection{Comparison Methods}
For comparison, we adopt a classical stereo matching method Semi-Global Matching (SGM) \cite{hirschmuller2007stereo} and several unsupervised methods that can be divided into two categories.
The first category includes three solutions using the photometric loss.
Besides the baseline unsupervised network trained under the setting of S1 (denoted as BaseNet) as mentioned in Sec.~\ref{sec_challenge}, we further use the state-of-the-art non-blind SR method RCAN \cite{zhang2018image} and blind SR method DAN \cite{luo2020unfolding} to super-resolve the LR view as pre-processing, denoted as RCAN+BaseNet and DAN+BaseNet, respectively.
The RCAN model is trained under the BIC degradation on a large-scale dataset DIV2K \cite{agustsson2017ntire} for SR, while the DAN model is trained under a set of degradations including BIC, IG, and AG on DIV2K.
The second category includes two feature-metric learning methods \cite{shu2020feature,spencer2020defeat} that also adopt the baseline network but impose feature-metric consistency in respective feature spaces as mentioned in Sec.~\ref{sec_feature_selec}, denoted as BaseNet+CL and BaseNet+AE, respectively.
Note that, unless SR models are used, bicubic interpolation is applied to upsample the LR view.
 
The backbone network of all learning-based solutions is the popular PSMNet \cite{chang2018pyramid}.
The network is optimized with the ADAM solver ($\beta_1$=$0.9$, $\beta_1$=$0.999$).
We set the learning rate as 0.001.
The smoothness constraint on disparity is enforced by the weighted smoothness loss \cite{li2018occlusion}, \emph{i.e.},
\vspace{-1.25mm}
\begin{equation}
   \mathcal{L}_{sm}=\left|\partial_{x} d_{L}\right| e^{-\left|\partial_{x} I_{L}\right|}+\left|\partial_{y} d_{L}\right| e^{-\left|\partial_{y} I_{L}\right|}.
   \vspace{-1.25mm}
\end{equation}
Therefore, the overall loss function of all learning-based solutions can be written as
\vspace{-1.5mm}
\begin{equation}
   \mathcal{L} = \mathcal{L}_{pm/fm} + \lambda\mathcal{L}_{sm},
   \vspace{-1.25mm}
\end{equation}
where $\lambda$ is a weighting factor and $\mathcal{L}_{pm/fm}$ is either the photometric loss for the methods in the first category or the corresponding feature-metric loss for the methods in the second category and ours.
The number of stages $K$ in the self-boosting strategy is set as 3.
The detailed architecture of the backbone network and the hyper-parameters of different methods are provided in the supplement.

\begin{figure*}
   \centering
   \includegraphics[width=1.005\linewidth]{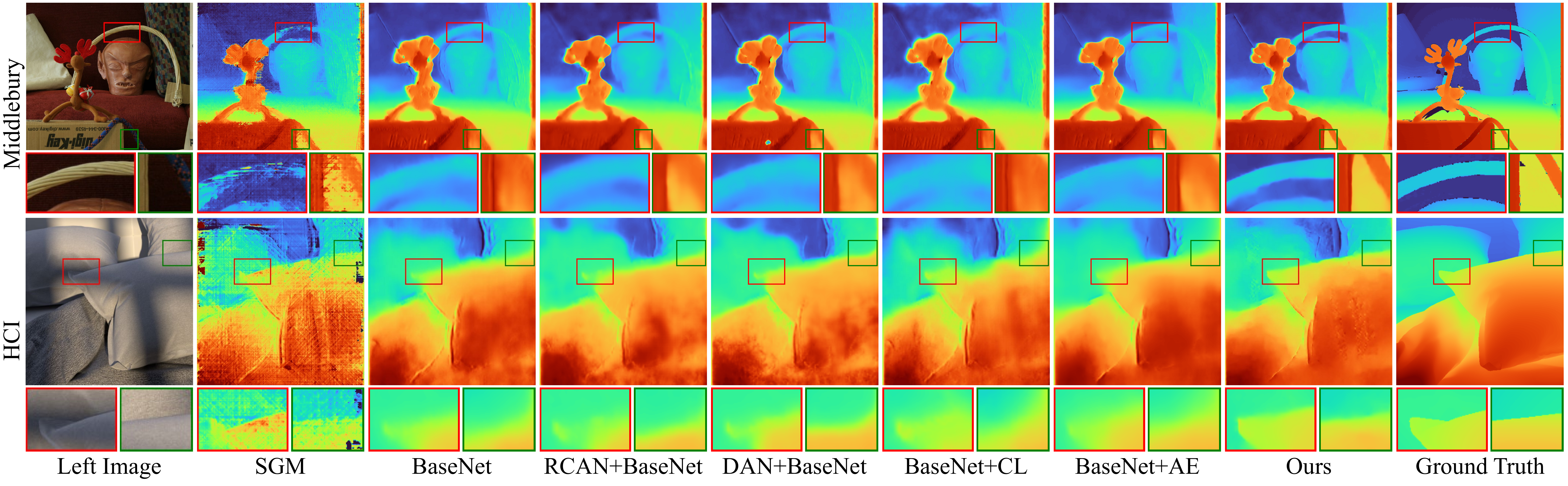}
   \vspace{-8mm}
   \caption{Disparity maps of two exemplar scenes from the Middlebury and HCI datasets.
   The first scene (\textit{Reindeer}) is simulated under the IG\_JPEG degradation, while the second scene (\textit{Pillows}) is simulated under the AG\_JPEG degradation.}
   \label{fig:visual_simulated}
   \vspace{-3.5mm}
 \end{figure*}

\subsection{Results}
\noindent \textbf{Quantitative Results.}
Table \ref{tab:overall_result} shows the comparison results of different methods on four simulated datasets with an asymmetric factor of 4.
Compared with methods that do not assume specific degradations (SGM, BaseNet, BaseNet+CL, and BaseNet+AE), our method has a distinct advantage on all datasets and under all degradations.
Although BaseNet+CL/AE also resorts to the feature-metric loss, the performance is only comparable or even inferior to BaseNet.
It tells that finding a degradation-agnostic and matching-specific feature space is non-trivial.
The comparison with the degradation-specific SR solutions RCAN+BaseNet and DAN+BaseNet should be interpreted in twofold.
On the one hand, when the actual degradations are \textit{consistent} with what they assume (BIC for RCAN and BIC/IG/AG for DAN), our method has better performance in most cases yet the improvement is not that large.
On the other hand, when the actual degradations are \textit{inconsistent} with their assumptions (marked gray in Table~\ref{tab:overall_result}), our method notably surpasses these SR solutions.
That is to say, SR solutions will lose efficacy when degradations are unknown in real-world scenarios.

\noindent \textbf{Visual Results.}
We provide the visual results of two exemplar scenes from the HCI and Middlebury datasets for comparison in Fig.~\ref{fig:visual_simulated}.
As can be seen, our method obtains more robust results, especially in regions with depth discontinuities.
In these regions, correspondence estimation is challenging for the solutions based on photometric consistency, since matching ambiguities could not be resolved even with the help of SR techniques.
In contrast, under the feature-metric consistency imposed in a degradation-agnostic and matching-specific feature space, our method better reveals the 3D geometry of testing scenes than BaseNet+CL/AE.

\begin{table}[!t]
   \caption{Comparison of different methods on datasets simulated with an asymmetric factor of 8 and under the BIC degradation.}
   \label{tab:large_scale}
   \vspace{-3mm}
   \setlength{\tabcolsep}{6pt}
   \renewcommand{\arraystretch}{1.125}
   \centering
   \resizebox{\columnwidth}{!}{
   \centering
   \begin{tabular}{ccccc}
      \toprule[1pt] 
      Method & Inria\_SLFD & HCI & Middlebury & KITTI2015 \\
      \hline
      SGM & 34.00/3.979 & 27.57/3.063 & 24.72/2.609 & 57.56/8.83 \\
      BaseNet & 27.93/2.963 & 23.21/2.164 & 15.33/2.049 & 38.88/4.673 \\
      RCAN+BaseNet & 21.17/2.442 & 11.54/1.331 & 11.28/1.729 & 25.92/3.159 \\
      BaseNet+CL & 32.49/3.337 & 15.16/1.589 & 16.51/2.129 & 53.28/5.571 \\
      BaseNet+AE & 27.11/2.847 & 12.13/1.450 & 14.30/2.020 & 30.81/3.299 \\
      Ours & \textbf{13.30}/\textbf{1.763} & \textbf{6.17}/\textbf{1.008} & \textbf{9.90}/\textbf{1.584} & \textbf{19.10}/\textbf{2.545} \\      
      \bottomrule[1pt] 
   \end{tabular}
   }
   \vspace{-4mm}
\end{table}

\noindent \textbf{Large Asymmetric Factor.}
To evaluate the performance of different methods\footnote{DAN \cite{luo2020unfolding} does not provide the model for scale 8 officially.} under large degradations, we conduct experiments on different datasets simulated with an asymmetric factor of 8 and under the BIC degradation.
As can be seen from Table~\ref{tab:large_scale}, our method surpasses all comparison methods by a large margin, and the improvement is even larger compared with the results in Table~\ref{tab:overall_result}.
For methods using the photometric loss, their performance deteriorates further due to the more severe photometric inconsistency.
In contrast, thanks to the self-boosting strategy, our method progressively strengthens the feature-metric consistency and thus maintains superior performance.

\begin{figure*}
   \centering
   \includegraphics[width=\linewidth]{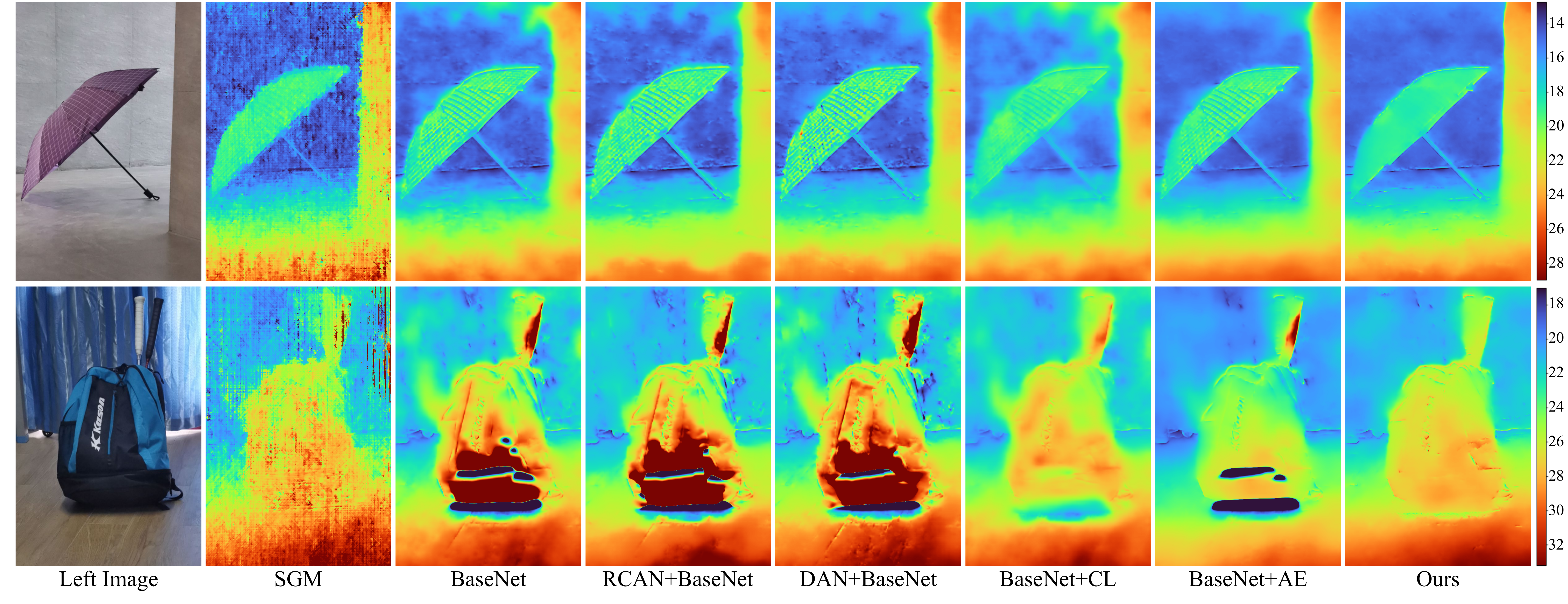}
   \vspace{-8.5mm}
   \caption{Disparity maps of two exemplar scenes from the self-collected real-world dataset. The colorbars show the value of disparity.}
   \label{fig:visual_realworld}
   \vspace{-2mm}
\end{figure*}

\begin{table}[!t]
   \caption{Comparison with supervised learning on the Middlebury and KITTI2015 datasets simulated with an asymmetric factor of 4 and under the IG degradation.}
   \label{tab:supervised}
   \vspace{-3mm}
   \setlength{\tabcolsep}{6pt}
   \renewcommand{\arraystretch}{1.1}
   \centering
   \resizebox{\columnwidth}{!}{
   \centering
   \begin{tabular}{ccccc}
      \toprule[1pt]
      Training & Testing & BaseNet-su & BaseNet & Ours \\
      \hline
      Middlebury & Middlebury & \textbf{4.05}/\textbf{0.906} & 9.50/1.482 & 6.52/1.178 \\
      Middlebury & KITTI2015 & 19.46/3.965 & 16.98/2.541 & \textbf{13.14}/\textbf{2.280} \\
      \bottomrule[1pt] 
   \end{tabular}
   }
   \vspace{-1.25mm}
\end{table}

\noindent \textbf{Comparison with Supervised Learning.}
The focus of this work is unsupervised learning that does not require ground-truth disparity labels during training and is more robust to be deployed in diverse real-world systems.
To verify this point, we also implement a supervised method, which uses the same backbone network as ours but leverages the ground-truth disparity to compute a smooth $L_1$ loss \cite{chang2018pyramid} (denoted as BaseNet-su).
We conduct experiments on the Middlebury and KITTI2015 datasets with an asymmetric factor of 4 under the IG degradation.
For both datasets, the networks are trained on Middlebury.
Since KITTI2015 consists of street scenes while Middlebury consists of indoor scenes, these two datasets have a large domain gap.
As shown in Table~\ref{tab:supervised}, when trained on Middlebury and tested on the same dataset, BaseNet-su has the best performance, which is reasonable.
However, when trained on Middlebury and tested on KITTI2015, the supervised method loses efficacy and our method achieves notably better generalization, demonstrating the robustness of our method in real-world scenarios where the disparity labels are not available for training.

\begin{table}[!t]
   \caption{Investigation on the backbone network. * indicates PSMNet \cite{chang2018pyramid} is replaced by iResNet \cite{liang2018learning}. Datasets are simulated with an asymmetric factor of 4 under the BIC degradation.}
   \label{tab:ablationg_model_archi}
   \vspace{-3mm}
   \setlength{\tabcolsep}{6pt}
   \renewcommand{\arraystretch}{1.125}
   \centering
   \resizebox{\columnwidth}{!}{
   \centering
   \begin{tabular}{ccccc}
      \toprule[1pt] 
      Method & Inria\_SLFD & HCI & Middlebury & KITTI2015 \\
      \hline
      BaseNet$^*$ & 18.80/2.411 & 18.58/1.964 & 10.92/1.769 & 17.82/2.549 \\ %
      Ours$^*$ & \textbf{9.83}/\textbf{1.407} & \textbf{5.83}/\textbf{0.866} & \textbf{8.39}/\textbf{1.382} & \textbf{10.86}/\textbf{1.960} \\ %
      \bottomrule[1pt] 
   \end{tabular}
   }
   \vspace{-3.75mm}
\end{table}

\noindent \textbf{Investigation on the Backbone Network.}
\label{sec:model_archi}
Besides PSMNet that adopts 3D convolution layers, we also investigate iResNet \cite{liang2018learning} as another embodiment of the backbone network of our method, which is purely based on 2D convolution layers. 
Experiments are conducted under the BIC degradation with an asymmetric factor of 4.
As shown in Table~\ref{tab:ablationg_model_archi}, the iResNet version of our method shows significant gains over the baseline network trained with the photometric loss on all datasets.
It demonstrates that the feature extractor of iResNet also learns degradation-agnostic and matching-specific features, which can be used to establish the feature-metric consistency.
In other words, the effectiveness of our method is independent of the backbone network used.


\section{Experiments on a Real-world Dataset}
\vspace{-1mm}
\label{sec_realworld}
\noindent \textbf{Dataset Preparation.}
To validate the performance of our method in real-world systems, we collect a resolution-asymmetric stereo dataset with real degradations.
The asymmetric stereo pairs are captured with a Huawei P30 smartphone.
This smartphone is equipped with a tele-wide camera system, including a 27mm-equivalent primary lens and an 80mm-equivalent tele-photo lens.
The asymmetric factor is approximately equal to 3.
After camera calibration and stereo rectification, we capture 30 asymmetric stereo pairs for indoor and outdoor scenes.
We randomly split 5 pairs as the testing set and the others as the training set.

\noindent \textbf{Results.}
As shown in Fig.~\ref{fig:visual_realworld}, our method achieves the best visual quality in comparison with the competitors.
Similar to the results on simulated datasets, our method estimates much sharper edges and better separates the objects belonging to different depth levels.
This advantage is essential to downstream applications, such as bokeh \cite{wadhwa2018synthetic} and 3D photography \cite{shih20203d}.
In contrast, methods using the photometric loss replicate some undesired textures from input images to the estimated disparity maps (\emph{e.g.}, the surface of the umbrella), which is mainly caused by the photometric inconsistency during stereo matching.
Both SR solutions show negligible improvements over the baseline network, since their degradation assumptions deviate from the real ones.
In addition, the other two methods using the feature-metric loss generate unsatisfactory results due to their incompetent feature spaces.
More results are provided in the supplement.

\section{Limitation and Conclusion}
\vspace{-1mm}
\noindent \textbf{Limitation.}
Besides resolution, there might exist other kinds of asymmetry (\emph{e.g.}, color and brightness) when capturing stereo images with a tele-wide camera system, due to the inherent optical differences of two lenses.
When collecting the real-world dataset, we manually adjust ISO, exposure time, and white balance of two lenses to mitigate these issues. 
Although they can be further alleviated through an explicit color and brightness correction after global registration, it remains an open problem whether other kinds of asymmetry can be directly addressed by extending the proposed method.
We will consider it as future work.

\noindent \textbf{Conclusion.}
In this paper, we reveal the main challenge of unsupervised correspondence estimation from resolution-asymmetric stereo images, \emph{i.e.}, the violation of photometric consistency.
To conquer this challenge, we realize the feature-metric consistency in an effective and efficient way and introduce a self-boosting strategy to strengthen this consistency.
As validated by comprehensive experiments, our method demonstrates superior performance in dealing with various degradations between two views in practice.

\section*{Acknowledgement}
\vspace{-1mm}
We acknowledge funding from National Key R\&D Program of China under Grant 2017YFA0700800, and National Natural Science Foundation of China under Grants 62131003, 62021001, 61901435, and U19B2038.

{\small
\bibliographystyle{ieee_fullname}
\bibliography{reference.bib}

\begin{thebibliography}{10}\itemsep=-1pt

\bibitem{agustsson2017ntire}
Eirikur Agustsson and Radu Timofte.
\newblock Ntire 2017 challenge on single image super-resolution: Dataset and
  study.
\newblock In {\em CVPR Workshops}, 2017.

\bibitem{aleotti2020reversing}
Filippo Aleotti, Fabio Tosi, Li Zhang, Matteo Poggi, and Stefano Mattoccia.
\newblock Reversing the cycle: self-supervised deep stereo through enhanced
  monocular distillation.
\newblock In {\em ECCV}, 2020.

\bibitem{bulat2018learn}
Adrian Bulat, Jing Yang, and Georgios Tzimiropoulos.
\newblock To learn image super-resolution, use a gan to learn how to do image
  degradation first.
\newblock In {\em ECCV}, 2018.

\bibitem{cai2019toward}
Jianrui Cai, Hui Zeng, Hongwei Yong, Zisheng Cao, and Lei Zhang.
\newblock Toward real-world single image super-resolution: A new benchmark and
  a new model.
\newblock In {\em CVPR}, 2019.

\bibitem{chang2018pyramid}
Jia-Ren Chang and Yong-Sheng Chen.
\newblock Pyramid stereo matching network.
\newblock In {\em CVPR}, 2018.

\bibitem{chen2019camera}
Chang Chen, Zhiwei Xiong, Xinmei Tian, Zheng-Jun Zha, and Feng Wu.
\newblock Camera lens super-resolution.
\newblock In {\em CVPR}, 2019.

\bibitem{chen2018image}
Jingwen Chen, Jiawei Chen, Hongyang Chao, and Ming Yang.
\newblock Image blind denoising with generative adversarial network based noise
  modeling.
\newblock In {\em CVPR}, 2018.

\bibitem{cheng2020hierarchical}
Xuelian Cheng, Yiran Zhong, Mehrtash Harandi, Yuchao Dai, Xiaojun Chang,
  Hongdong Li, Tom Drummond, and Zongyuan Ge.
\newblock Hierarchical neural architecture search for deep stereo matching.
\newblock In {\em NeurIPS}, 2020.

\bibitem{Cheng_2021_CVPR}
Zhen Cheng, Zhiwei Xiong, Chang Chen, Dong Liu, and Zheng-Jun Zha.
\newblock Light field super-resolution with zero-shot learning.
\newblock In {\em CVPR}, 2021.

\bibitem{dong2014learning}
Chao Dong, Chen~Change Loy, Kaiming He, and Xiaoou Tang.
\newblock Learning a deep convolutional network for image super-resolution.
\newblock In {\em ECCV}, 2014.

\bibitem{egnal2000mutual}
Geoffrey Egnal.
\newblock Mutual information as a stereo correspondence measure.
\newblock 2000.

\bibitem{heo2010robust}
Yong~Seok Heo, Kyong~Mu Lee, and Sang~Uk Lee.
\newblock Robust stereo matching using adaptive normalized cross-correlation.
\newblock {\em IEEE Transactions on Pattern Analysis and Machine Intelligence},
  33(4):807--822, 2010.

\bibitem{hirschmuller2007stereo}
Heiko Hirschmuller.
\newblock Stereo processing by semiglobal matching and mutual information.
\newblock {\em IEEE Transactions on Pattern Analysis and Machine Intelligence},
  30(2):328--341, 2007.

\bibitem{hirschmuller2007evaluation}
Heiko Hirschmuller and Daniel Scharstein.
\newblock Evaluation of cost functions for stereo matching.
\newblock In {\em CVPR}, 2007.

\bibitem{hirschmuller2008evaluation}
Heiko Hirschmuller and Daniel Scharstein.
\newblock Evaluation of stereo matching costs on images with radiometric
  differences.
\newblock {\em IEEE Transactions on Pattern Analysis and Machine Intelligence},
  31(9):1582--1599, 2008.

\bibitem{honauer2016dataset}
Katrin Honauer, Ole Johannsen, Daniel Kondermann, and Bastian Goldluecke.
\newblock A dataset and evaluation methodology for depth estimation on 4d light
  fields.
\newblock In {\em ACCV}, 2016.

\bibitem{huang2020real}
Yukun Huang, Zheng-Jun Zha, Xueyang Fu, Richang Hong, and Liang Li.
\newblock Real-world person re-identification via degradation invariance
  learning.
\newblock In {\em CVPR}, 2020.

\bibitem{kendall2017end}
Alex Kendall, Hayk Martirosyan, Saumitro Dasgupta, Peter Henry, Ryan Kennedy,
  Abraham Bachrach, and Adam Bry.
\newblock End-to-end learning of geometry and context for deep stereo
  regression.
\newblock In {\em ICCV}, 2017.

\bibitem{li2018occlusion}
Ang Li and Zejian Yuan.
\newblock Occlusion aware stereo matching via cooperative unsupervised
  learning.
\newblock In {\em ACCV}, 2018.

\bibitem{Li_VCIP_2021}
Yue Li, Yueyi Zhang, and Zhiwei Xiong.
\newblock Revisiting flipping strategy for learning-based stereo depth
  estimation.
\newblock In {\em VCIP}, 2021.

\bibitem{liang2019unsupervised}
Mingyang Liang, Xiaoyang Guo, Hongsheng Li, Xiaogang Wang, and You Song.
\newblock Unsupervised cross-spectral stereo matching by learning to
  synthesize.
\newblock In {\em AAAI}, 2019.

\bibitem{liang2018learning}
Zhengfa Liang, Yiliu Feng, Yulan Guo, Hengzhu Liu, Wei Chen, Linbo Qiao, Li
  Zhou, and Jianfeng Zhang.
\newblock Learning for disparity estimation through feature constancy.
\newblock In {\em CVPR}, 2018.

\bibitem{liu2021blind}
Anran Liu, Yihao Liu, Jinjin Gu, Yu Qiao, and Chao Dong.
\newblock Blind image super-resolution: A survey and beyond.
\newblock {\em arXiv preprint arXiv:2107.03055}, 2021.

\bibitem{liu2020stereogan}
Rui Liu, Chengxi Yang, Wenxiu Sun, Xiaogang Wang, and Hongsheng Li.
\newblock Stereogan: Bridging synthetic-to-real domain gap by joint
  optimization of domain translation and stereo matching.
\newblock In {\em CVPR}, 2020.

\bibitem{liu2020visually}
Yicun Liu, Jimmy Ren, Jiawei Zhang, Jianbo Liu, and Mude Lin.
\newblock Visually imbalanced stereo matching.
\newblock In {\em CVPR}, 2020.

\bibitem{luo2020unfolding}
Zhengxiong Luo, Yan Huang, Shang Li, Liang Wang, and Tieniu Tan.
\newblock Unfolding the alternating optimization for blind super resolution.
\newblock In {\em NeurIPS}, 2020.

\bibitem{mayer2016large}
Nikolaus Mayer, Eddy Ilg, Philip Hausser, Philipp Fischer, Daniel Cremers,
  Alexey Dosovitskiy, and Thomas Brox.
\newblock A large dataset to train convolutional networks for disparity,
  optical flow, and scene flow estimation.
\newblock In {\em CVPR}, 2016.

\bibitem{menze2015object}
Moritz Menze and Andreas Geiger.
\newblock Object scene flow for autonomous vehicles.
\newblock In {\em CVPR}, 2015.

\bibitem{pan2016continuous}
Kuang-Yu Pan and Yung-Yu Chuang.
\newblock Continuous zoom with two fixed-focal-length lens.
\newblock In {\em SIGGRAPH ASIA 2016 Posters}.

\bibitem{peng2020zero}
Jiayong Peng, Zhiwei Xiong, Yicheng Wang, Yueyi Zhang, and Dong Liu.
\newblock Zero-shot depth estimation from light field using a convolutional
  neural network.
\newblock {\em IEEE Transactions on Computational Imaging}, 6:682--696, 2020.

\bibitem{scharstein2002taxonomy}
Daniel Scharstein and Richard Szeliski.
\newblock A taxonomy and evaluation of dense two-frame stereo correspondence
  algorithms.
\newblock {\em International journal of computer vision}, 47(1):7--42, 2002.

\bibitem{shi2019framework}
Jinglei Shi, Xiaoran Jiang, and Christine Guillemot.
\newblock A framework for learning depth from a flexible subset of dense and
  sparse light field views.
\newblock {\em IEEE Transactions on Image Processing}, 28(12):5867--5880, 2019.

\bibitem{shih20203d}
Meng-Li Shih, Shih-Yang Su, Johannes Kopf, and Jia-Bin Huang.
\newblock 3d photography using context-aware layered depth inpainting.
\newblock In {\em CVPR}, 2020.

\bibitem{shu2020feature}
Chang Shu, Kun Yu, Zhixiang Duan, and Kuiyuan Yang.
\newblock Feature-metric loss for self-supervised learning of depth and
  egomotion.
\newblock In {\em ECCV}, 2020.

\bibitem{spencer2020defeat}
Jaime Spencer, Richard Bowden, and Simon Hadfield.
\newblock Defeat-net: general monocular depth via simultaneous unsupervised
  representation learning.
\newblock In {\em CVPR}, 2020.

\bibitem{tonioni2019real}
Alessio Tonioni, Fabio Tosi, Matteo Poggi, Stefano Mattoccia, and Luigi~Di
  Stefano.
\newblock Real-time self-adaptive deep stereo.
\newblock In {\em CVPR}, 2019.

\bibitem{trinidad2019multi}
Marc~Comino Trinidad, Ricardo~Martin Brualla, Florian Kainz, and Janne
  Kontkanen.
\newblock Multi-view image fusion.
\newblock In {\em ICCV}, 2019.

\bibitem{wadhwa2018synthetic}
Neal Wadhwa, Rahul Garg, David~E Jacobs, Bryan~E Feldman, Nori Kanazawa, Robert
  Carroll, Yair Movshovitz-Attias, Jonathan~T Barron, Yael Pritch, and Marc
  Levoy.
\newblock Synthetic depth-of-field with a single-camera mobile phone.
\newblock {\em ACM Transactions on Graphics}, 37(4):1--13, 2018.

\bibitem{wang2020parallax}
Longguang Wang, Yulan Guo, Yingqian Wang, Zhengfa Liang, Zaiping Lin, Jungang
  Yang, and Wei An.
\newblock Parallax attention for unsupervised stereo correspondence learning.
\newblock {\em IEEE Transactions on Pattern Analysis and Machine Intelligence},
  2020.

\bibitem{Wang_2018_TCSVT}
Lizhi Wang, Zhiwei Xiong, Guangming Shi, Wenjun Zeng, and Feng Wu.
\newblock Simultaneous depth and spectral imaging with a cross-modal stereo
  system.
\newblock {\em IEEE Transactions on Circuits and Systems for Video Technology},
  28(3):812--817, 2018.

\bibitem{wang2021dual}
Tengfei Wang, Jiaxin Xie, Wenxiu Sun, Qiong Yan, and Qifeng Chen.
\newblock Dual-camera super-resolution with aligned attention modules.
\newblock In {\em ICCV}, 2021.

\bibitem{wang2020deep}
Yang Wang, Yang Cao, Zheng-Jun Zha, Jing Zhang, and Zhiwei Xiong.
\newblock Deep degradation prior for low-quality image classification.
\newblock In {\em CVPR}, 2020.

\bibitem{wang2021asymmetric}
Yicheng Wang, Jiayong Peng, Yueyi Zhang, Shan Liu, Xiaoyan Sun, and Zhiwei
  Xiong.
\newblock Asymmetric stereo color transfer.
\newblock In {\em ICME}, 2021.

\bibitem{xu2010two}
Li Xu and Jiaya Jia.
\newblock Two-phase kernel estimation for robust motion deblurring.
\newblock In {\em ECCV}, 2010.

\bibitem{yao2019spectral}
Mingde Yao, Zhiwei Xiong, Lizhi Wang, Dong Liu, and Xuejin Chen.
\newblock Spectral-depth imaging with deep learning based reconstruction.
\newblock {\em Optics express}, 27(26):38312--38325, 2019.

\bibitem{zhan2018unsupervised}
Huangying Zhan, Ravi Garg, Chamara~Saroj Weerasekera, Kejie Li, Harsh Agarwal,
  and Ian Reid.
\newblock Unsupervised learning of monocular depth estimation and visual
  odometry with deep feature reconstruction.
\newblock In {\em CVPR}, 2018.

\bibitem{zhang2019zoom}
Xuaner Zhang, Qifeng Chen, Ren Ng, and Vladlen Koltun.
\newblock Zoom to learn, learn to zoom.
\newblock In {\em CVPR}, 2019.

\bibitem{zhang2018image}
Yulun Zhang, Kunpeng Li, Kai Li, Lichen Wang, Bineng Zhong, and Yun Fu.
\newblock Image super-resolution using very deep residual channel attention
  networks.
\newblock In {\em ECCV}, 2018.

\bibitem{zhi2018deep}
Tiancheng Zhi, Bernardo~R Pires, Martial Hebert, and Srinivasa~G Narasimhan.
\newblock Deep material-aware cross-spectral stereo matching.
\newblock In {\em CVPR}, 2018.

\bibitem{zhong2017self}
Yiran Zhong, Yuchao Dai, and Hongdong Li.
\newblock Self-supervised learning for stereo matching with self-improving
  ability.
\newblock {\em arXiv preprint arXiv:1709.00930}, 2017.

\bibitem{zhou2017unsupervised}
Chao Zhou, Hong Zhang, Xiaoyong Shen, and Jiaya Jia.
\newblock Unsupervised learning of stereo matching.
\newblock In {\em ICCV}, 2017.

\bibitem{zhou2019fast}
Shenglong Zhou, Zhiwei Xiong, Chang Chen, Xuejin Chen, Dong Liu, Yueyi Zhang,
  Zheng-Jun Zha, and Feng Wu.
\newblock Fast and accurate electron microscopy image registration with 3d
  convolution.
\newblock In {\em MICCAI}, 2019.

\end{thebibliography}
}

\end{document}